\documentclass[10pt,twocolumn,letterpaper]{article}

\usepackage[pagenumbers]{wacv}      

\definecolor{wacvblue}{rgb}{0.21,0.49,0.74}
\usepackage[pagebackref,breaklinks,colorlinks,allcolors=wacvblue]{hyperref}
\usepackage{amsmath, amsfonts, amssymb, amsthm}

\theoremstyle{plain}
\newtheorem{theorem}{Theorem}
\newtheorem{lemma}{Lemma}
\newtheorem{corollary}{Corollary}
\newtheorem{proposition}{Proposition}

\theoremstyle{definition}
\newtheorem{assumption}{Assumption}

\def\wacvPaperID{2654} 
\def\confName{WACV}
\def\confYear{2027}

\title{Localization Is Sufficient for GPS-Denied Aerial Navigation}

\author{Rukai Gao\\
Washington University in St.\ Louis
\and
Michael Lanier\\
Washington University in St.\ Louis
\and
Eric Xing\\
Washington University in St.\ Louis
\and
Nathan Jacobs\\
Washington University in St.\ Louis
}

\begin{document}

\maketitle
\begin{abstract}
Zero-shot visual place recognition against georeferenced aerial imagery supports accurate GPS-denied UAV position estimation without task-specific training, but existing results rely on prescribed flight paths, inertial odometry, and maps that describe the current scene. We study what happens when the loop is closed: a UAV released at an unknown position must reach a known goal using only sequential observations and a pre-acquired reference map, with no route, no inertial odometry in the localization chain, and no training. We formalize this setting as a continuous POMDP and prove a finite-sample bound showing that the belief concentrates at a rate given by the difference between two divergences, terrain distinctiveness and reference-map staleness, and that the controller enters this rate only by choosing where the margin is sampled. A matching lower bound follows: where the margin is near-uniform over the reachable region, an information-seeking detour can improve concentration by at most a map-determined factor, and where staleness dominates it cannot help at all. Between these lies a regime of sharply varying margin in which a detour can pay, and which our benchmarks do not isolate. Empirically, we find that localization is the binding constraint: a particle filter weighted by off-the-shelf foundation-model descriptors and guided by direct goal-pointing reaches success rates of 0.91 and 0.60 across mild and severe map-change benchmarks (temporally extended SwissView and post-disaster xBD), outperforming the strongest reinforcement learning baseline (0.76 and 0.36) and a SLAM+$A^*$ pipeline (0.51 and 0.08). The advantage widens as the map ages: SLAM+$A^*$ falls from 0.51 to 0.08 under structural change, the belief-space approach from 0.91 to 0.60. As the analysis predicts, four active-localization mechanisms give no improvement on these two benchmarks and often cost efficiency.
\end{abstract}    
\section{Introduction}
\label{sec:intro}

\begin{figure*}[t]
    \centering
    \includegraphics[width=0.70\textwidth]{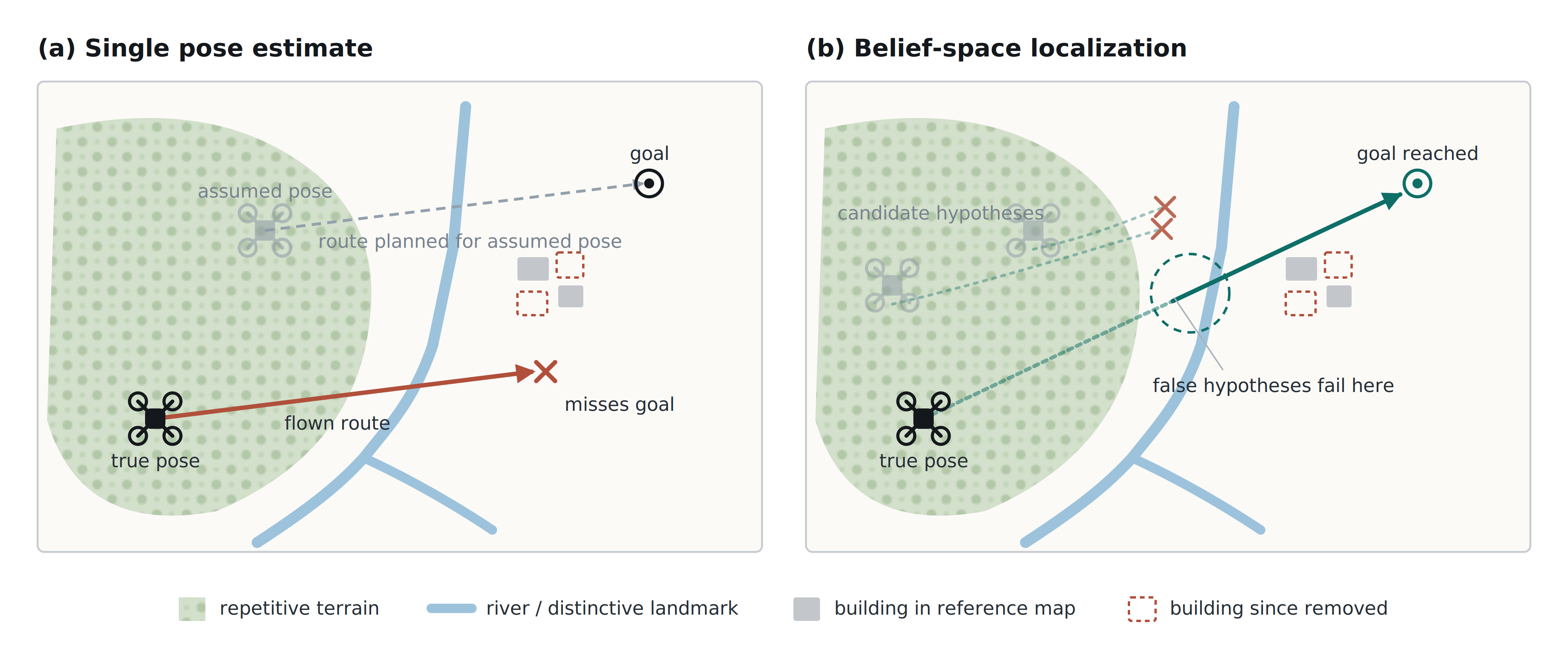}
    \caption{\textbf{Navigating against a reference map that is no longer current.} The UAV is released at an unknown position and must reach a known goal. Over visually repetitive terrain, an observation is consistent with many map locations at once. \textbf{(a)} A method restricted to a single pose estimate must commit to one of them. The route it plans is correct for the pose it assumes, but executing that route from the true pose lands the UAV elsewhere, and nothing in the system registers the error. \textbf{(b)} Retaining competing hypotheses instead, the UAV continues toward the goal without committing, eliminating false hypotheses as it passes terrain whose appearance differs under each of them.}
    \label{fig:teaser}
\end{figure*}

Autonomous UAV flight generally assumes GPS for global positioning. That assumption breaks in disaster response over damaged infrastructure, contested airspace, and urban canyons with severe multipath. The standard alternative is to localize visually against georeferenced aerial imagery, and the approach has matured: particle filters over satellite imagery recover global pose from uninformed starts~\cite{mantelli2019novel,kinnari2023lsvl}, and foundation-model place descriptors match zero-shot across seasons, lighting, and weather~\cite{keetha2023anyloc,foundloc2023}. However, existing evaluations remain open-loop: pose is measured along prescribed routes with inertial odometry anchoring estimates between fixes, and location estimates do not steer the vehicle.

Closing the control loop changes the problem in two ways. First, errors propagate dynamically: mislocalization steers the UAV off-course, drawing subsequent observations from unintended locations. Second, reference maps are historical. Vegetation shifts, construction alters terrain, and disaster-response settings contain destroyed anchor structures by definition. Appearance-invariant matching addresses visual variance, but cannot handle missing structural anchors.

We study the setting where a UAV is released at an unknown position and must reach a known goal from sequential camera observations alone, against a stale reference map, with no route, no fine-tuning, and no inertial odometry. The aircraft still carries an IMU, which any airframe needs for attitude stabilization; what we withhold is inertial odometry as an input to localization, so that the visual channel can be evaluated on its own. Single-estimate SLAM pipelines drift when visual correspondences weaken, executing erroneous plans with undiminished confidence (Fig.~\ref{fig:teaser}(a)). Learned policies internalize appearance variations but require per-environment training and lack explicit uncertainty representations.

Representing ambiguity avoids premature commitment over repetitive terrain (Fig.~\ref{fig:teaser}(b)). Much like a human navigating with an outdated map, heading directly toward the target lets ambiguity resolve as distinctive landmarks are encountered on the way; detours help only when the nominal path has no distinguishing features, and not at all when those features are missing from the map.

We argue that for goal-directed flight in open terrain, localization is the binding constraint. How fast an accurate belief forms is set by map distinctiveness and staleness, which the route can sample but not change. Sec.~\ref{sec:theory} makes this precise: the belief concentrates at a rate equal to the gap between terrain distinctiveness and map staleness, and the controller chooses only where that gap is sampled. A particle filter weighted by off-the-shelf descriptors and guided by a naive goal-pointing controller achieves success rates of 0.91 and 0.60 on temporally extended SwissView100~\cite{Mi_2025_ICCV} and post-disaster xBD~\cite{Gupta_2019_CVPR_Workshops}, outperforming RL (0.76/0.36) and SLAM+$A^*$ (0.51/0.08) baselines. Four standard active-localization mechanisms give no gain on either benchmark, because on both the margin varies little over the region a detour could reach. Our contributions are:
\begin{itemize}
    \item \textbf{Benchmark to be publicly released for reference-map staleness.} We construct two calibrated evaluation regimes isolating temporal change (SwissView100) and severe structural modification (xBD), and report baselines for zero-shot GPS-denied control under map drift. We will release the benchmark with all code, gym environments, and artifacts.
    \item \textbf{Theoretical bounds on passive control sufficiency.} We formalize the problem as a continuous POMDP and give a finite-sample bound (Thm.~\ref{thm:sufficiency}) on how fast the belief concentrates under a passive goal-pointing controller. The rate is the gap between terrain distinctiveness and map staleness, both divergences fixed by the map. A matching lower bound (Prop.~\ref{prop:converse}) shows that no policy concentrates the belief faster. Together they cap what an active controller can gain (Cor.~\ref{cor:regimes}) by how much the margin varies over the reachable region, and they explain the negative result below.
    \item \textbf{Sufficiency of belief-space localization.} We show empirically that passive belief maintenance outperforms learned policies and SLAM pipelines under map degradation, and that on both benchmarks active motion strategies do not improve on direct goal-directed flight.
\end{itemize}
\section{Related Work}
\label{sec:related}

\noindent\textbf{Absolute Visual Localization.}
Global UAV positioning without GPS typically pairs map matching with particle filters or factor graphs~\cite{couturier2021review,goforth2019gps,mantelli2019novel,kinnari2023lsvl}, with earlier work registering onboard views against public orthoimagery~\cite{patel2020visual}. These pipelines evaluate pose open-loop along prescribed routes, with inertial odometry anchoring estimates between visual fixes. We withhold inertial odometry from the localization chain and close the control loop, so errors compound into the flight path.

\noindent\textbf{Zero-Shot Place Recognition.}
Foundation-model descriptors enable condition-invariant place recognition without fine-tuning~\cite{keetha2023anyloc}. Recent frameworks like FoundLoc~\cite{foundloc2023} combine zero-shot descriptors with inertial odometry for open-loop pose estimation. We test whether zero-shot features alone are sufficient to \emph{drive} goal-directed flight, and we evaluate them under severe structural map staleness as well as photometric and seasonal variation.

\noindent\textbf{Active Localization \& Belief Planning.}
Active localization and POMDP belief-space planning select trajectories to minimize pose uncertainty under Monte Carlo sampling~\cite{burgard1997active,thrun2005probabilistic,vandenberg2012motion}. These methods assume reference maps faithfully reflect the world. Convergence guarantees for Monte Carlo localization likewise presume a map matching the environment~\cite{dellaert1999monte,thrun2005probabilistic,fox2003adapting}, and cover the filter in isolation but not a vehicle steered by it. \cite{lanier2024psrl} considers planning where state information is present in hindsight. 
We give a closed-loop bound under a map known to be stale, whose rate is set by the map--world discrepancy, with the action sequence entering only through where the margin is sampled. When maps are structurally out of date, ambiguity comes from map inaccuracy and exploration cannot recover information the map never contained.

\noindent\textbf{Learned Aerial Navigation.}
Active geo-localization uses reinforcement learning to search for targets from known start poses~\cite{pirinen2023aerial,sarkar2024gomaa,Mi_2025_ICCV}, requiring per-environment training and lacking explicit uncertainty representations. We invert the setting (known goal, unknown start) and show that zero-shot belief filtering outperforms learned policies with no training.
\section{Problem Statement}
\label{sec:problem_statement}

We consider GPS-denied aerial navigation where a UAV starts at an unknown position $p_0 \in \Omega$ and must navigate to a known goal $p_g \in \Omega$ using only a downward-facing camera and a pre-acquired reference map $\mathcal{M}$. No GPS is available, and no inertial or visual odometry feeds the localization chain, so dead reckoning is unavailable; the IMU an airframe needs for attitude stabilization is assumed present but is not used to estimate position. We estimate position in the map plane at fixed altitude and nadir attitude, not full 6-DoF pose, so the observation footprint is determined by position alone. The map $\mathcal{M}$ is also temporally stale: visual content during deployment differs from reference imagery because of seasonal, structural, and environmental changes.

At step $t$, the UAV occupies true position $p_t \in \Omega$, executes continuous motion $a_t \in \mathcal{A}$, and captures local observation $z_t$. Realized state transitions follow $p_{t+1} \sim T(p_{t+1} \mid p_t, a_t, e_t)$ under actuation uncertainty and environmental disturbances $e_t$. Given history $h_t = (z_0, a_0, \dots, a_{t-1}, z_t)$, the UAV maintains a posterior belief over its current position $b_t(p) = P(p_t = p \mid h_t, \mathcal{M})$, initialized uniformly over $\Omega$ and approximated by a weighted particle set:
\begin{equation}
    \mathcal{P}_t = \left\{ \left(p_t^{(n)}, w_t^{(n)}\right) \right\}_{n=1}^{N}, \qquad \sum_{n=1}^{N} w_t^{(n)} = 1.
    \label{eq:particles}
\end{equation}

We maintain the full distribution $\mathcal{P}_t$ and not a point estimate, because repetitive terrain makes $P(z_t \mid p_t, \mathcal{M})$ multimodal and map staleness miscalibrates it systematically. Sec.~\ref{sec:theory} states when deferring pose commitment is guaranteed to dissipate the false hypotheses, and shows that the rate is set by the map, not by the route.
\section{Method}
\label{sec:method}

We evaluate whether zero-shot belief-space localization, closed into a navigation loop, is sufficient to reach a goal under reference-map staleness. The pipeline is deliberately minimal: a particle filter over the mapped region, weighted by off-the-shelf foundation-model descriptors matched against the reference map, paired with a controller that steers toward the goal from the current belief estimate. Sec.~\ref{sec:extensions} describes four additional mechanisms, drawn from the active-localization literature, that we implement and evaluate; their effect is reported in Sec.~\ref{sec:ablation}.

\subsection{Reference-Map Localization}
\label{sec:localization}

\noindent\textbf{Offline map description.}\\*
We partition the geo-referenced reference map $\mathcal{M}$ into crops with known centers $\{x_i\}_{i=1}^{L}$, laid on a grid of stride half the crop width, so neighboring crops overlap by half, and extract an AnyLoc~\cite{keetha2023anyloc} descriptor $f_i$ for each crop. AnyLoc is used off the shelf, with no fine-tuning on either benchmark, so the pipeline requires no task-specific training data. This computation depends only on $\mathcal{M}$ and is performed once per map, prior to deployment.

\noindent\textbf{Observation likelihood.}\\*
For an incoming observation $z_t$, its descriptor is compared with all reference descriptors,
\begin{equation}
    s_t(j)=\operatorname{sim}(f(z_t),f_j).
    \label{eq:similarity}
\end{equation}
The visual likelihood is a Boltzmann distribution over this similarity field,
\begin{equation}
    L_t(j)
    =
    \exp
    \left\{
        \gamma
        \left[s_t(j)-\max_k s_t(k)\right]
    \right\},
    \label{eq:likelihood}
\end{equation}
which attains its maximum at the best-matching crop and decays with similarity deficit. The inverse temperature $\gamma$ is fixed across observations and benchmarks.

\noindent\textbf{Particle filter.}\\*
The UAV maintains the weighted particle belief $\mathcal{P}_t$ of Eq.~\eqref{eq:particles}. Since the initial position is unknown, particles are initialized approximately uniformly over $\Omega$. Given a commanded heading $\theta_t$ and magnitude $m_t$, each particle is propagated under stochastic heading and magnitude errors together with persistent environmental disturbance,
\begin{equation}
    p_{t+1}^{(n)}
    =
    p_t^{(n)}
    +
    m_t^{(n)}
    \begin{bmatrix}
        \cos\theta_t^{(n)}\\
        \sin\theta_t^{(n)}
    \end{bmatrix}
    +e_t^{(n)},
    \label{eq:motion}
\end{equation}
and weights are updated by
\begin{equation}
    w_t^{(n)}
    \propto
    w_{t^-}^{(n)}
    L_t(c(p_t^{(n)})),
    \label{eq:weights}
\end{equation}
where $c(\cdot)$ returns the index of the reference crop containing a given position. Because the likelihood in Eq.~\eqref{eq:likelihood} is flat wherever the similarity field is flat, an observation consistent with many map locations produces a correspondingly diffuse weight update and leaves competing hypotheses intact, without any explicit mechanism for detecting that this has occurred.

\subsection{Goal-Directed Control}
\label{sec:control}

The current location estimate is the weighted particle mean,
\begin{equation}
    \hat p_t = \sum_n w_t^{(n)} p_t^{(n)},
    \label{eq:mean}
\end{equation}
and the controller points from this estimate toward the known goal,
\begin{equation}
    \theta_t^{\mathrm{goal}}
    =
    \operatorname{arctan2}
    (p_{g,y}-\hat p_{t,y},
     p_{g,x}-\hat p_{t,x}).
    \label{eq:heading}
\end{equation}
The UAV flies toward the goal under whatever pose estimate the belief currently supports, and the trajectory is corrected implicitly as the belief concentrates. A heading is a setpoint that an inner attitude and velocity loop tracks. That loop runs orders of magnitude faster, on inertial feedback, and holds the aircraft up. The guidance layer is intentionally as simple as possible. If the belief is accurate, reaching the goal requires nothing further, and localization carries the whole task.

\subsection{Evaluated Extensions}
\label{sec:extensions}

The pipeline above makes no attempt to detect ambiguity or to act on it, and its controller does not deviate from the goal heading to improve localization. Each of the four mechanisms below addresses a case where the plain pipeline may fall short: it sharpens the likelihood, restores lost hypotheses, summarizes a multimodal belief, or steers toward terrain where observations discriminate more strongly. Each has precedent in the active-localization literature and augments or replaces one stage of the pipeline. Sec.~\ref{sec:ablation} measures their contribution.

\noindent\textbf{Map-conditioned observation utility.}\\*
A reference location is difficult to localize when it is visually similar to geographically distant regions. We define its ambiguity as
\begin{equation}
    A_i=
    \sum_j
    \left[\max(0,s_{ij}-\tau_s)\right]^p
    \left[
        1-\exp\left(-\frac{d_{ij}^2}{d_0^2}\right)
    \right],
    \label{eq:ambiguity}
\end{equation}
where $s_{ij}=f_i^\top f_j$, $d_{ij}=\|x_i-x_j\|_2$, and $d_0$ sets the separation beyond which a lookalike is counted. The second factor suppresses nearby lookalikes, whose confusion costs little, and retains distant ones, whose confusion is costly. Normalizing $A_i$ gives a utility $U_i\in[0,1]$, and the same construction on the observation similarity field yields $U_t$; replacing $\gamma$ in Eq.~\eqref{eq:likelihood} with $\gamma U_t$ then lets distinctive observations sharpen the belief while ambiguous ones barely move it.

\noindent\textbf{Hypothesis reinvigoration.}\\*
A finite particle population cannot recover a hypothesis once every particle representing it has disappeared, but replacing many particles on the strength of one observation destabilizes localization under appearance change. When a high-utility observation disagrees strongly with the predicted belief, measured by $G_t=\sum_n w_{t^-}^{(n)} L_t(c(p_t^{(n)}))/\max_j L_t(j)$, we sample a small proposal set from the global likelihood. Proposals are held outside the operational filter and promoted only if their evidence exceeds that of the main belief by a prescribed ratio after the next observation; otherwise they are discarded.

\noindent\textbf{Robust point estimate.}\\*
Because a multimodal belief need not have a meaningful arithmetic mean, Eq.~\eqref{eq:mean} can place the estimate between modes, in a location the posterior considers implausible. The weighted geometric median $\operatorname{arg\,min}_p \sum_n w_t^{(n)}\|p_t^{(n)}-p\|_2$ remains at or near a mode instead.

\noindent\textbf{Belief-aware recovery.}\\*
Belief spread is measured by $R_{90}$, the minimum radius around the estimate containing $90\%$ of the posterior mass. Multimodality is identified by fitting Gaussian mixture models with varying component counts and selecting among them by the integrated classification likelihood, subject to minimum component mass and spatial separation. When either broad spread or multimodality persists across several observations, the controller departs from Eq.~\eqref{eq:heading} in favor of an information-seeking maneuver.

Candidates are relative displacements in a goal-facing frame, so that one candidate can be scored under every hypothesis the belief holds: by utility alone when the belief is merely broad, and by utility weighted with the expected descriptor disagreement between mode pairs when it is multimodal, since modes are separated only by evidence that differs between them. Beam search ranks candidate trajectories by recovery value per unit of extra travel, the highest-ranked are verified by Monte Carlo rollout of vehicle and filter together, and only the first action is executed before the decision is reconsidered. The supplementary material gives the scoring functions in full.
\section{Theoretical Analysis}
\label{sec:theory}

\begin{figure*}[t]
    \centering
    \includegraphics[width=\textwidth]{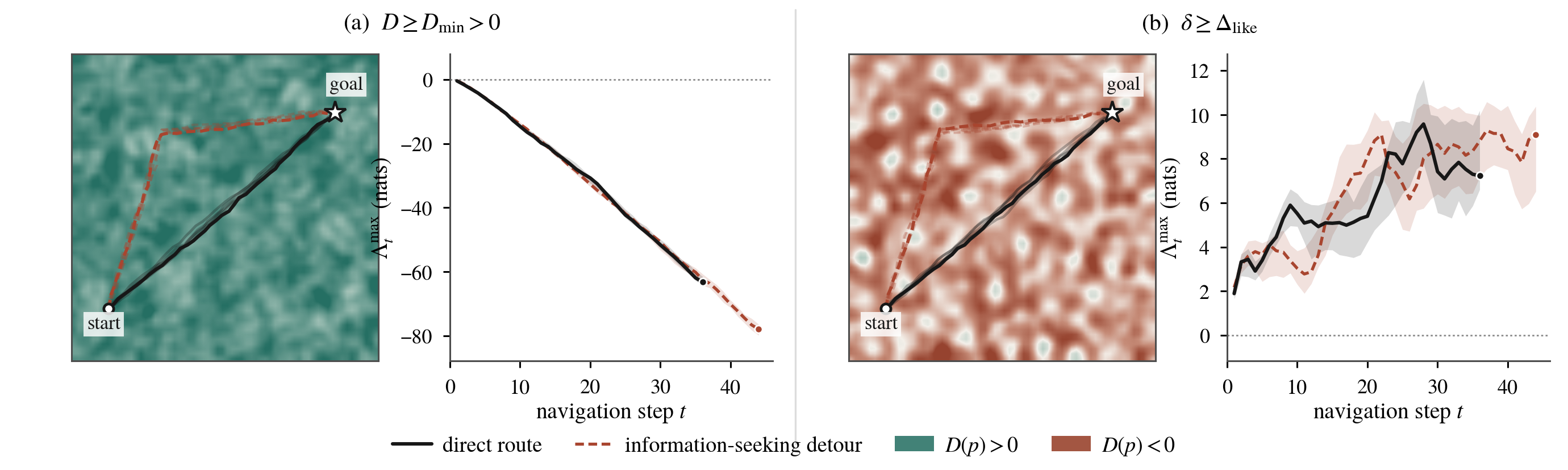}
    \caption{\textbf{The two regimes of Cor.~\ref{cor:regimes}, on a controlled instance of the model analyzed here.} The map either still describes the world \textbf{(a)} or has been structurally overwritten \textbf{(b)}, and the filter of Sec.~\ref{sec:method} is flown along two \emph{prescribed} routes, direct and information-seeking, which therefore differ only in where the margin $D$ of Eq.~\eqref{eq:discrim} is sampled. Left of each pair: the operating region shaded by $D$ at $\|\Delta\|=\rho$. Right: median and interquartile range over 24 episodes of $\Lambda_t^{\max}=\max_{\|\Delta\|\ge\rho}\log w_t(p_t{+}\Delta)/w_t(p_t)$, the quantity Thm.~\ref{thm:sufficiency} bounds, ending at the goal. \textbf{(a)} With the margin positive, both routes accumulate evidence at the same rate and localize by $t=6$; the detour buys nothing for its $20\%$ of extra path. \textbf{(b)} With staleness dominating, $\Lambda_t^{\max}$ stays above zero and neither route localizes: the evidence a detour would gather is missing from the map, not merely unvisited.}
    \label{fig:regimes}
\end{figure*}

We now establish when the pipeline of Sec.~\ref{sec:method}, without its extensions, suffices to reach the goal. The analysis isolates one per-step quantity governing how fast the belief concentrates, shows that it is a property of map and terrain and not of the trajectory, and characterizes two regimes in which no controller improves it.

\noindent\textbf{Setup.} Sec.~\ref{sec:problem_statement} gives the ingredients of the POMDP $\langle \mathcal{S}, \mathcal{A}, T, \mathcal{Z}, O_{\mathcal{M}}, R \rangle$ with $\mathcal{S}=\Omega$, heading actions, transitions from Eq.~\eqref{eq:motion}, and observations scored against $\mathcal{M}$ by $O_{\mathcal{M}}$ from Eq.~\eqref{eq:likelihood}. Deployment observations come from the \emph{world} $O_{\mathcal{W}}(\cdot\mid p)$, which the map no longer matches; the belief of Eq.~\eqref{eq:particles} is updated by $b_{t+1}\propto b_t\,O_{\mathcal{M}}$, and the policy is Eq.~\eqref{eq:heading} acting on the mean of Eq.~\eqref{eq:mean}. We analyze these importance weights directly; resampling is a variance-reduction step that leaves their expectations unchanged.

\begin{assumption}[Bounded evidence per step]
\label{assump:bounded}
There is $B<\infty$ with $\big|\log O_{\mathcal{M}}(z\mid p) - \log O_{\mathcal{M}}(z\mid q)\big| \le B$ for all $z$ and all $p,q\in\Omega$.
\end{assumption}

\noindent This holds by construction here. Reading Eq.~\eqref{eq:likelihood} as a density in $z$ with normalizer $Z(p)$ gives $\log O_{\mathcal{M}}(z\mid p)-\log O_{\mathcal{M}}(z\mid q) = \gamma\,(s_t(c(p))-s_t(c(q)))-\log Z(p)/Z(q)$, so with cosine similarities in $[-1,1]$, $B\le 2\gamma+\omega$ for $\omega$ the oscillation of $\log Z$ over $\Omega$, both fixed before deployment.

\noindent\textbf{Per-step discriminability.} For a true position $p$ and a hypothesis displaced by $\Delta$, define
\begin{equation}
    D(p,\Delta) \;\triangleq\; \mathbb{E}_{z\sim O_{\mathcal{W}}(\cdot\mid p)}\!\left[\log\frac{O_{\mathcal{M}}(z\mid p)}{O_{\mathcal{M}}(z\mid p+\Delta)}\right],
    \label{eq:discrim}
\end{equation}
the expected log-evidence that one observation at $p$ contributes against that hypothesis.

\begin{assumption}[Predictable evidence]
\label{assump:condind}
Conditioned on the history, the expected log-evidence one observation contributes against a displaced hypothesis is $D(p_\tau,\Delta)$: $\mathbb{E}[\ell_\tau\mid\mathcal{F}_{\tau-1}]=-D(p_\tau,\Delta)$.
\end{assumption}

\noindent Conditional independence of observations given the true positions, the usual particle-filtering assumption, implies this but is stronger than needed: only the conditional mean enters the martingale argument below. Overlapping fields of view and a reused descriptor therefore do not break the bound: the correlation they introduce reduces the evidence a step adds beyond what the history already supplies, shrinking $D_{\min}$ and lengthening $T_0$.

\begin{lemma}[Terrain information decomposition]
\label{lem:decomp}
Let $\Delta_{\mathrm{like}}(p,\Delta)=\mathrm{KL}\big(O_{\mathcal{W}}(\cdot\mid p)\,\|\,O_{\mathcal{M}}(\cdot\mid p+\Delta)\big)$ and $\delta(p)=\mathrm{KL}\big(O_{\mathcal{W}}(\cdot\mid p)\,\|\,O_{\mathcal{M}}(\cdot\mid p)\big)$. Then
\begin{equation}
    D(p,\Delta) \;=\; \Delta_{\mathrm{like}}(p,\Delta) \;-\; \delta(p),
    \label{eq:decomp}
\end{equation}
so $D(p,\Delta)>0$ if and only if $\Delta_{\mathrm{like}}(p,\Delta)>\delta(p)$.
\end{lemma}
\begin{proof}
Write $\mathbb{E}_{z\sim O_{\mathcal{W}}(\cdot\mid p)}[\log O_{\mathcal{M}}(z\mid q)] = \mathbb{E}[\log O_{\mathcal{W}}(z\mid p)] - \mathrm{KL}(O_{\mathcal{W}}(\cdot\mid p)\|O_{\mathcal{M}}(\cdot\mid q))$ and apply it to $q=p$ and $q=p+\Delta$ in Eq.~\eqref{eq:discrim}; the world entropy cancels.
\end{proof}

\noindent $\delta(p)$ measures how poorly the stale map explains the world at the correct location and $\Delta_{\mathrm{like}}$ how poorly it explains it at a displaced one, so an observation is informative exactly when distinctiveness exceeds staleness.

\begin{lemma}[Offset invariance]
\label{lem:offset}
Let $\Delta_t$ be the displacement between a hypothesis and the true position, and let the filter propagate particles under the vehicle's own error distribution, of per-step covariance $\Sigma$. Then for every action sequence $\{\theta_\tau\}_{\tau\le t}$: \emph{(i)} $\mathbb{E}[\Delta_t]=\Delta_0$, and \emph{(ii)} $\operatorname{tr}\operatorname{Cov}(\Delta_t)=\operatorname{tr}\operatorname{Cov}(\Delta_0)+t\operatorname{tr}\Sigma$. Only when the per-step error is isotropic does the rotation drop out of the recursion, making the whole distribution of $\Delta_t$, and not just these two moments, action-independent.
\end{lemma}
\begin{proof}
The controller of Eq.~\eqref{eq:heading} is a function of the belief, not of any hypothesis, so it issues one command $(\theta_t,m_t)$ that every particle and the vehicle receive alike. Differencing two instances of Eq.~\eqref{eq:motion} under it, the commanded displacement cancels and leaves $\Delta_{t+1}=\Delta_t+R(\theta_t)\eta_t$, with $R$ a rotation and $\eta_t$ the difference between the vehicle's error draw and the particle's. By hypothesis the two come from the same distribution and are independent of $\mathcal{F}_t$, so $\eta_t$ has mean zero; $\theta_t$ is $\mathcal{F}_t$-measurable, and taking expectations gives (i). For (ii), $\operatorname{Cov}(\Delta_t)=\operatorname{Cov}(\Delta_0)+\sum_{\tau<t}R(\theta_\tau)\Sigma R(\theta_\tau)^\top$, and each summand has trace $\operatorname{tr}\Sigma$ whatever $\theta_\tau$ is. Actions rotate the covariance but cannot shrink it: $\Delta_t$ is a driftless random walk whose spread grows as $t\operatorname{tr}\Sigma$, absorbed into $\rho$.
\end{proof}

\noindent A controller therefore cannot alter $\Delta$, nor $D(\cdot,\cdot)$, which the map and the world fix. It influences the belief through one channel only: which positions $p_\tau$ are visited, and hence where $D$ is sampled.
\begin{theorem}[Finite-sample concentration]
\label{thm:sufficiency}
Under Assumptions~\ref{assump:bounded}--\ref{assump:condind}, for any $\eta$ s.t. $0 < \eta <1$ and any horizon $T$, with probability at least $1-\eta$ the following holds simultaneously for all $N$ particles and all $t\le T$:
\begin{equation}
\begin{split}
    \log\frac{w_t(p')}{w_t(p_t)} \;\le\;& \log\frac{w_0(p')}{w_0(p_0)} \;-\; \sum_{\tau=1}^{t} D(p_\tau,\Delta)\\
    &+\; 2B\sqrt{2t\log(NT/\eta)}.
\end{split}
    \label{eq:concentration}
\end{equation}
If in addition the margin of Lemma~\ref{lem:decomp} is uniformly positive along the realized trajectory, $\Delta_{\mathrm{like}}(p_\tau,\Delta)-\delta(p_\tau)\ge D_{\min}>0$ for every $\|\Delta\|\ge\rho$, then every hypothesis outside $\mathcal{B}_\rho(p_t)$ has vanishing relative weight once
\begin{equation}
    t \;\ge\; T_0 \;=\; \mathcal{O}\!\left(\frac{B^2\log(NT/\eta)}{D_{\min}^2}\right),
    \label{eq:T0}
\end{equation}
after which $\hat p_t$ tracks $p_t$ to within $\rho$ and, in an obstacle-free domain, the controller reaches $\mathcal{B}_\epsilon(p_g)$, $\epsilon\ge\rho$, in finite time.
\end{theorem}

\begin{proof}
The normalizer is common to all particles, so it cancels in the ratio and $\log\frac{w_t(p')}{w_t(p_t)}=\log\frac{w_0(p')}{w_0(p_0)}+\Lambda_t$ with $\Lambda_t=\sum_{\tau\le t}\ell_\tau$ and $\ell_\tau=\log\frac{O_{\mathcal{M}}(z_\tau\mid p_\tau+\Delta)}{O_{\mathcal{M}}(z_\tau\mid p_\tau)}$. By Assumption~\ref{assump:condind} and Eq.~\eqref{eq:discrim}, $\mathbb{E}[\ell_\tau\mid\mathcal{F}_{\tau-1}]=-D(p_\tau,\Delta)$, so $\Lambda_t+\sum_{\tau\le t}D(p_\tau,\Delta)$ is a martingale whose increments are bounded by $2B$ under Assumption~\ref{assump:bounded}. Azuma--Hoeffding bounds its deviation by $2B\sqrt{2t\log(1/\eta')}$ with probability $1-\eta'$; a union bound over the $N$ particles and $T$ steps with $\eta'=\eta/(NT)$ gives Eq.~\eqref{eq:concentration}, uniformly in $t\le T$ rather than at a single time. Being uniform over $\|\Delta\|\ge\rho$, the margin also covers the displacement that Lem.~\ref{lem:offset} lets wander, while it stays outside $\mathcal{B}_\rho$. Under the uniform margin the drift term is at least $tD_{\min}$, which dominates the $\sqrt{t}$ fluctuation once $t$ satisfies Eq.~\eqref{eq:T0}, driving the ratio to zero. The posterior then concentrates in $\mathcal{B}_\rho(p_t)$, so $\hat p_t$ is accurate to within $\rho$, the heading error in Eq.~\eqref{eq:heading} is at most $\arcsin(\rho/\|p_g-\hat p_t\|)<\pi/2$, and each step reduces $\|p_g-p_t\|$ in expectation while that distance exceeds $\rho$.
\end{proof}

\noindent The guarantee is silent before $T_0$. While the belief is multimodal the mean of Eq.~\eqref{eq:mean} can fall between modes, so the heading is unconstrained and the vehicle may travel as far as $T_0$ steps off course. Without obstacles this costs path length alone, which is why the plain mean suffices in Eq.~\eqref{eq:heading}; where collisions are possible it does not, and a mode-consistent estimate such as the geometric median of Sec.~\ref{sec:extensions}, or an explicit safety constraint, is required.

\begin{proposition}[No policy concentrates faster than the map allows]
\label{prop:converse}
Suppose $D(p,\Delta)\le D_{\max}$ for all $p\in\Omega_{\mathrm{op}}$ and all $\|\Delta\|\ge\rho$. Fix $\varepsilon,\eta\in(0,1)$. Then under \emph{every} admissible policy, with probability at least $1-\eta$, no hypothesis at displacement $\Delta$ is driven below relative weight $\varepsilon$ before step
\begin{equation}
  \tau_\varepsilon \;\ge\; \min\!\left\{\frac{\log(1/\varepsilon)}{2D_{\max}},\;\frac{\log^2(1/\varepsilon)}{32B^2\log(1/\eta)}\right\}.
  \label{eq:converse}
\end{equation}
\end{proposition}

\begin{proof}
Where Thm.~\ref{thm:sufficiency} lower-bounds the drift to show that concentration is fast enough, here we upper-bound it to show that no policy makes concentration arbitrarily fast. With $\Lambda_t$ and the martingale $M_t$ of that proof, $\Lambda_t=M_t-\sum_{\tau\le t}D(p_\tau,\Delta)\ge M_t-tD_{\max}$, since no admissible policy leaves $\Omega_{\mathrm{op}}$ and $D\le D_{\max}$ there. Azuma--Hoeffding on the lower tail of $M_t$, whose increments are bounded by $2B$ under Assumption~\ref{assump:bounded}, gives $M_t\ge-2B\sqrt{2t\log(1/\eta)}$ with probability at least $1-\eta$, so
\begin{equation*}
  \Lambda_t \;\ge\; -tD_{\max} \;-\; 2B\sqrt{2t\log(1/\eta)} .
\end{equation*}
Driving the hypothesis below relative weight $\varepsilon$ requires $\Lambda_t\le-\log(1/\varepsilon)$, hence $tD_{\max}+2B\sqrt{2t\log(1/\eta)}\ge\log(1/\varepsilon)$. For $t<\log(1/\varepsilon)/(2D_{\max})$ the drift term is below $\tfrac12\log(1/\varepsilon)$, and for $t<\log^2(1/\varepsilon)/(32B^2\log(1/\eta))$ the fluctuation term is; below the minimum of the two both hold, so their sum cannot reach $\log(1/\varepsilon)$ and no such $t$ exists.
\end{proof}

\begin{corollary}[Two regimes in which motion cannot help]
\label{cor:regimes}
Let $\Omega_{\mathrm{op}}$ be the region reachable within the travel budget, let $\kappa$ be the extra path length a detour spends over direct flight, write $D_{\max}=\sup_{\Omega_{\mathrm{op}}}D$, and read $D=\Delta_{\mathrm{like}}-\delta$ via Lemma~\ref{lem:decomp}. Both conditions below are required for all $p\in\Omega_{\mathrm{op}}$ and all $\|\Delta\|\ge\rho$. \emph{(i)} If $D_{\min}\le D$ with $D_{\min}>0$, every admissible policy, the passive one included, concentrates within $T_0$ of Eq.~\eqref{eq:T0}, and by Prop.~\ref{prop:converse} none concentrates before Eq.~\eqref{eq:converse}. Dividing the two, the speedup a detour can buy over passive flight is at most a factor $\mathcal{O}\!\left(B^2D_{\max}\log(NT/\eta)\,/\,D_{\min}^2\log(1/\varepsilon)\right)$, a constant fixed by the map and not by the controller, and the detour pays $\kappa$ on top of it. What governs that constant is the heterogeneity $D_{\max}/D_{\min}$ of the margin over $\Omega_{\mathrm{op}}$: where the margin is near-uniform a route has nothing to exploit, and where it varies sharply (open water beside a feature-rich coastline) a detour may be worth its length. \emph{(ii)} If $\delta\ge\Delta_{\mathrm{like}}$, then $D\le0$ by Lemma~\ref{lem:decomp} and the policy cannot move $\Delta$ by Lemma~\ref{lem:offset}, so $\mathbb{E}[\Lambda_t]=-\sum_\tau D(p_\tau,\Delta)\ge0$ for every reachable trajectory. No policy drives a false hypothesis down in expectation, because the evidence a detour would gather is absent from the map and not merely unvisited.
\end{corollary}

Fig.~\ref{fig:regimes} instantiates both regimes on a map whose margin is known by construction. Between them lies a third case the corollary does not cover, and which the heterogeneity term above identifies: a direct route that is uninformative while some reachable detour is not. There a detour can reduce $T_0$, and nothing here says otherwise. The bounds do not say which case real maps fall into, that is, whether terrain distinctiveness ($\Delta_{\mathrm{like}}$) outpaces staleness ($\delta$) on outdated imagery, and how often the two separate across a single reachable region. Sec.~\ref{sec:experiments} measures the success and cost of passive navigation against SLAM and RL baselines at two calibrated levels of staleness, and Sec.~\ref{sec:ablation} tests Cor.~\ref{cor:regimes} on the two regimes it covers.

\section{Experiments and Results}
\label{sec:experiments}

\paragraph{Baselines.}
We compare against three navigation baselines.
(i) \textbf{Double DQN}~\cite{vanhasselt2016deep} is a discrete-action reinforcement learning baseline that selects among eight motion headings using visual localization features and goal information.
(ii) \textbf{DDPG}~\cite{lillicrap2016continuous} uses the same navigation-state representation while predicting a continuous motion heading.
(iii) \textbf{SLAM+A$^{*}$} combines visual localization and odometry with pose-graph optimization and A$^{*}$ planning~\cite{grisetti2010tutorial,hart1968formal}: the initial pose is estimated globally from the first observation, after which SIFT visual odometry between consecutive frames and absolute registration against the reference map enter a pose graph as relative and absolute factors, and the single MAP pose drives the planner. It is map-anchored rather than pure odometry, so it does not drift in the usual unbounded sense while the map still describes the scene. It fails here because the absolute factors are what degrade: where structures are gone, registration returns a confident but wrong constraint, and with a single pose hypothesis there is nothing to contradict it. Full settings are in the supplementary material.
The reinforcement learning baselines are trained per environment; SLAM+A$^{*}$ and our method require no training. All methods share the same GPS-denied setting.

\paragraph{Evaluation metrics.}
We report success rate (SR), Success weighted by Path Length (SPL)~\cite{anderson2018evaluation}, and Success weighted by Completion Time (SCT)~\cite{yokoyama2021sct}. An episode succeeds when the UAV reaches the goal region without leaving the valid map or exceeding twice the straight-line distance to it.

For $N$ evaluation episodes, SPL is defined as
\begin{equation}
    \mathrm{SPL}
    =
    \frac{1}{N}
    \sum_{i=1}^{N}
    S_i
    \frac{l_i}{\max(p_i,l_i)},
    \label{eq:spl}
\end{equation}
where $S_i\in\{0,1\}$ denotes success, $l_i$ is the shortest start-to-goal distance, and $p_i$ is the path length traveled. SCT analogously measures success-weighted completion-time efficiency,
\begin{equation}
    \mathrm{SCT}
    =
    \frac{1}{N}
    \sum_{i=1}^{N}
    S_i
    \frac{T_i}{\max(C_i,T_i)},
    \label{eq:sct}
\end{equation}
where $C_i$ is the realized completion time and $T_i$ the minimum completion time under the same vehicle dynamics. Higher is better for all three; SPL and SCT are success rates discounted by path and time efficiency, respectively.

\noindent\textbf{Datasets.}\\*
We evaluate on two aerial-image benchmarks representing different levels of reference-to-deployment appearance change. Our primary benchmark is a temporal extension of SwissView100~\cite{Mi_2025_ICCV}, built from Swisstopo's SWISSIMAGE 10\,cm imagery: 100 scenes sampled across Switzerland spanning urban, rural, mountainous, agricultural, and natural terrain, each a $10{,}000\times10{,}000$ image covering $1\,\mathrm{km}\times1\,\mathrm{km}$.

\begin{figure}[t]
    \centering
    \includegraphics[width=\linewidth]{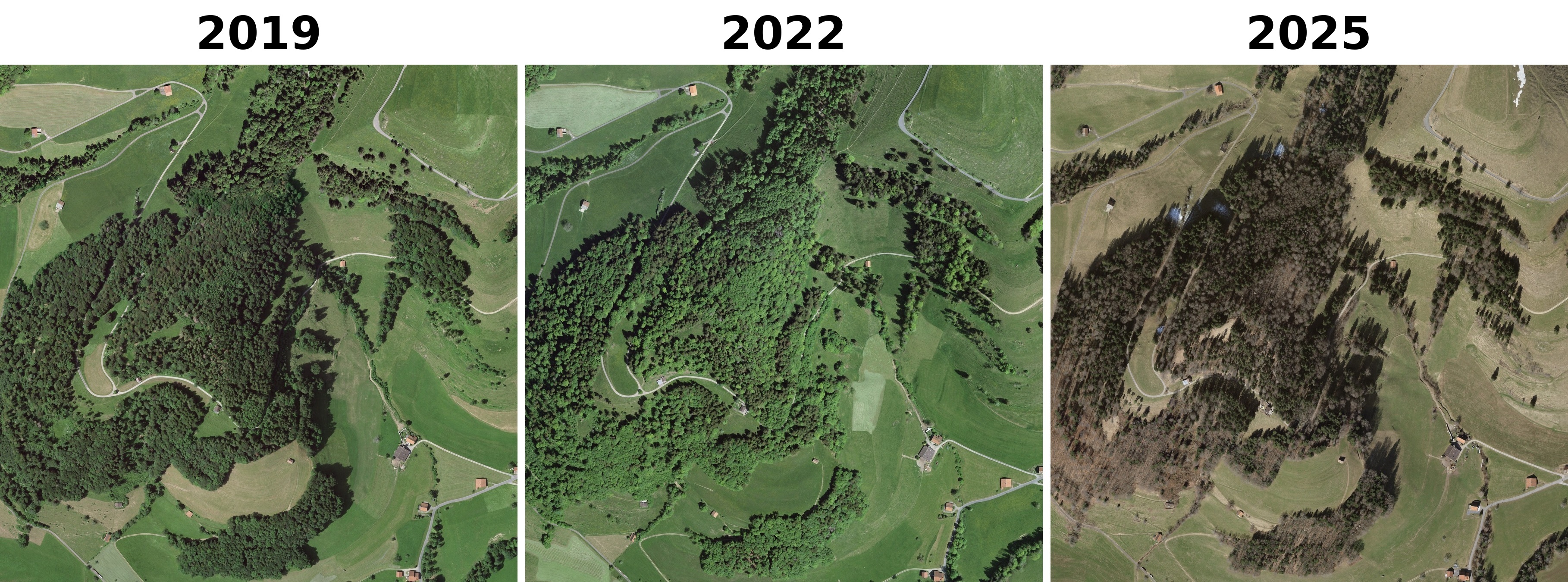}
    \caption{Example temporal acquisitions from the same SwissView scene
    in 2019, 2022, and 2025. The images cover the same
    $1\,\mathrm{km}^2$ geographic region and are pixel-aligned, while
    exhibiting temporal appearance changes across acquisitions.}
    \label{fig:temporal_swissview_example}
\end{figure}

We extend it with multiple acquisitions of the same 100 scenes, typically about three years apart: 44 scenes have two, 55 have three, and one has four, giving 257 registered images and, since the direction of temporal change is retained, 430 ordered reference--observation pairs. Acquisitions of a scene are pixel-aligned, so the benchmark isolates temporal appearance variation (vegetation, season, illumination, shadow, land use, construction) without artificial image transformations. Fig.~\ref{fig:temporal_swissview_example} shows one scene in 2019, 2022, and 2025.

Our second benchmark is xBD~\cite{Gupta_2019_CVPR_Workshops}, which provides paired pre- and post-disaster satellite imagery. Temporal SwissView varies mainly in appearance over time. xBD contains much stronger reference-to-deployment change caused by natural disasters, including large-scale structural modification and the disappearance or alteration of previously visible landmarks.

\noindent\textbf{Implementation details.}\\*
SwissView uses $512\times512$ local observations with a 256-pixel navigation step; xBD uses $128\times128$ observations with a 64-pixel step, and motion and disturbance magnitudes are scaled by $1/4$ accordingly. The descriptor is AnyLoc~\cite{keetha2023anyloc} used off the shelf, with no fine-tuning on either benchmark. Additional configuration and the reinforcement learning training details are in the supplementary material.

\noindent\textbf{Navigation performance.}\\*
Tables~\ref{tab:swissview-results} and~\ref{tab:xbd-results} report performance on the two benchmarks as mean $\pm$ half-width of the 95\% bootstrap confidence interval.

\begin{table}[t]
    \centering
    \caption{Navigation performance on the temporally extended SwissView test set (mild map staleness). Values are mean $\pm$ 95\% confidence interval; higher is better.}
    \label{tab:swissview-results}
    \resizebox{\columnwidth}{!}{
    \begin{tabular}{lccc}
        \toprule
        Method & SR $\uparrow$ & SPL $\uparrow$ & SCT $\uparrow$ \\
        \midrule
        Double DQN~\cite{vanhasselt2016deep}
        & $0.74 \pm 0.07$ & $0.67 \pm 0.07$ & $0.67 \pm 0.07$ \\
        DDPG~\cite{lillicrap2016continuous}
        & $0.76 \pm 0.07$ & $0.69 \pm 0.07$ & $0.69 \pm 0.07$ \\
        SLAM+A$^{*}$
        & $0.51 \pm 0.08$ & $0.49 \pm 0.08$ & $0.49 \pm 0.08$ \\
        \textbf{Belief-space (ours)}
        & $\mathbf{0.91 \pm 0.05}$ & $\mathbf{0.88 \pm 0.05}$ & $\mathbf{0.86 \pm 0.05}$ \\
        \bottomrule
    \end{tabular}}
\end{table}
\begin{table}[t]
    \centering
    \caption{Navigation performance on the xBD test set under severe pre-/post-disaster change (severe map staleness). Values are mean $\pm$ 95\% confidence interval; higher is better.}
    \label{tab:xbd-results}
    \resizebox{\columnwidth}{!}{
    \begin{tabular}{lccc}
        \toprule
        Method & SR $\uparrow$ & SPL $\uparrow$ & SCT $\uparrow$ \\
        \midrule
        Double DQN~\cite{vanhasselt2016deep}
        & $0.36 \pm 0.07$ & $0.32 \pm 0.06$ & $0.32 \pm 0.06$ \\
        DDPG~\cite{lillicrap2016continuous}
        & $0.34 \pm 0.07$ & $0.30 \pm 0.06$ & $0.30 \pm 0.06$ \\
        SLAM+A$^{*}$
        & $0.08 \pm 0.04$ & $0.07 \pm 0.03$ & $0.07 \pm 0.03$ \\
        \textbf{Belief-space (ours)}
        & $\mathbf{0.60 \pm 0.07}$ & $\mathbf{0.57 \pm 0.07}$ & $\mathbf{0.56 \pm 0.07}$ \\
        \bottomrule
    \end{tabular}}
\end{table}
\begin{table*}[t]
    \centering
    \caption{Ablation results on SwissView and xBD (mean $\pm$ 95\% CI); best per column in bold. No configuration separates from the bare pipeline beyond its confidence interval on success rate.}
    \label{tab:ablation_results}
    \resizebox{\textwidth}{!}{
    \begin{tabular}{lcccccc}
        \toprule
        & \multicolumn{3}{c}{SwissView (mild staleness)} & \multicolumn{3}{c}{xBD (severe staleness)} \\
        \cmidrule(lr){2-4} \cmidrule(lr){5-7}
        Configuration & SR $\uparrow$ & SPL $\uparrow$ & SCT $\uparrow$ & SR $\uparrow$ & SPL $\uparrow$ & SCT $\uparrow$ \\
        \midrule
        Full (all mechanisms) & $0.87 \pm 0.06$ & $0.77 \pm 0.05$ & $0.69 \pm 0.06$ & $\mathbf{0.60 \pm 0.07}$ & $0.54 \pm 0.06$ & $0.50 \pm 0.06$ \\
        $-$ Recovery & $\mathbf{0.92 \pm 0.04}$ & $\mathbf{0.88 \pm 0.04}$ & $\mathbf{0.86 \pm 0.04}$ & $0.55 \pm 0.07$ & $0.53 \pm 0.07$ & $0.52 \pm 0.07$ \\
        $-$ Recovery, $-$ Reinvig. & $\mathbf{0.92 \pm 0.04}$ & $\mathbf{0.88 \pm 0.04}$ & $\mathbf{0.86 \pm 0.04}$ & $\mathbf{0.60 \pm 0.07}$ & $\mathbf{0.57 \pm 0.07}$ & $\mathbf{0.56 \pm 0.07}$ \\
        $-$ Recovery, $-$ Reinvig., $-$ GeoMed & $0.90 \pm 0.05$ & $0.85 \pm 0.05$ & $0.83 \pm 0.05$ & $0.59 \pm 0.07$ & $0.57 \pm 0.07$ & $0.56 \pm 0.07$ \\
        Bare pipeline & $0.91 \pm 0.05$ & $\mathbf{0.88 \pm 0.05}$ & $\mathbf{0.86 \pm 0.05}$ & $\mathbf{0.60 \pm 0.07}$ & $\mathbf{0.57 \pm 0.07}$ & $\mathbf{0.56 \pm 0.07}$ \\
        \bottomrule
    \end{tabular}}
\end{table*}

On SwissView, belief-space localization reaches an SR of $0.91$ against $0.76$ for the strongest learned baseline and $0.51$ for SLAM+A$^{*}$, with the highest SPL and SCT as well. Efficiency is high in absolute terms ($\mathrm{SPL}=0.88$, $\mathrm{SCT}=0.86$): successful episodes follow near-direct routes, as expected when an accurate belief leaves little to correct.

On xBD the same approach retains an SR of $0.60$, against $0.36$ for the strongest learned baseline and $0.08$ for SLAM+A$^{*}$. The margin over the learned baselines widens from $0.15$ on SwissView to $0.24$ on xBD, and the gap over SLAM+A$^{*}$ widens from $0.40$ to $0.52$. The value of representing the full posterior therefore grows as the reference map becomes less faithful.

SLAM+A$^{*}$ is the most severely affected by structural change, falling from $0.51$ to $0.08$. It commits to a single pose from visual correspondences with the reference imagery; once structures are removed those correspondences remain and mislead, the pose graph absorbs the bad absolute factor, and the error propagates into the plan undetected. Retaining a distribution avoids this collapse: an observation consistent with several locations leaves the belief diffuse where a single pose would be confidently wrong, and the belief reconcentrates once the trajectory reaches terrain the map still describes. 

\begin{figure*}[!tb]
    \centering
    \includegraphics[width=0.85\textwidth]{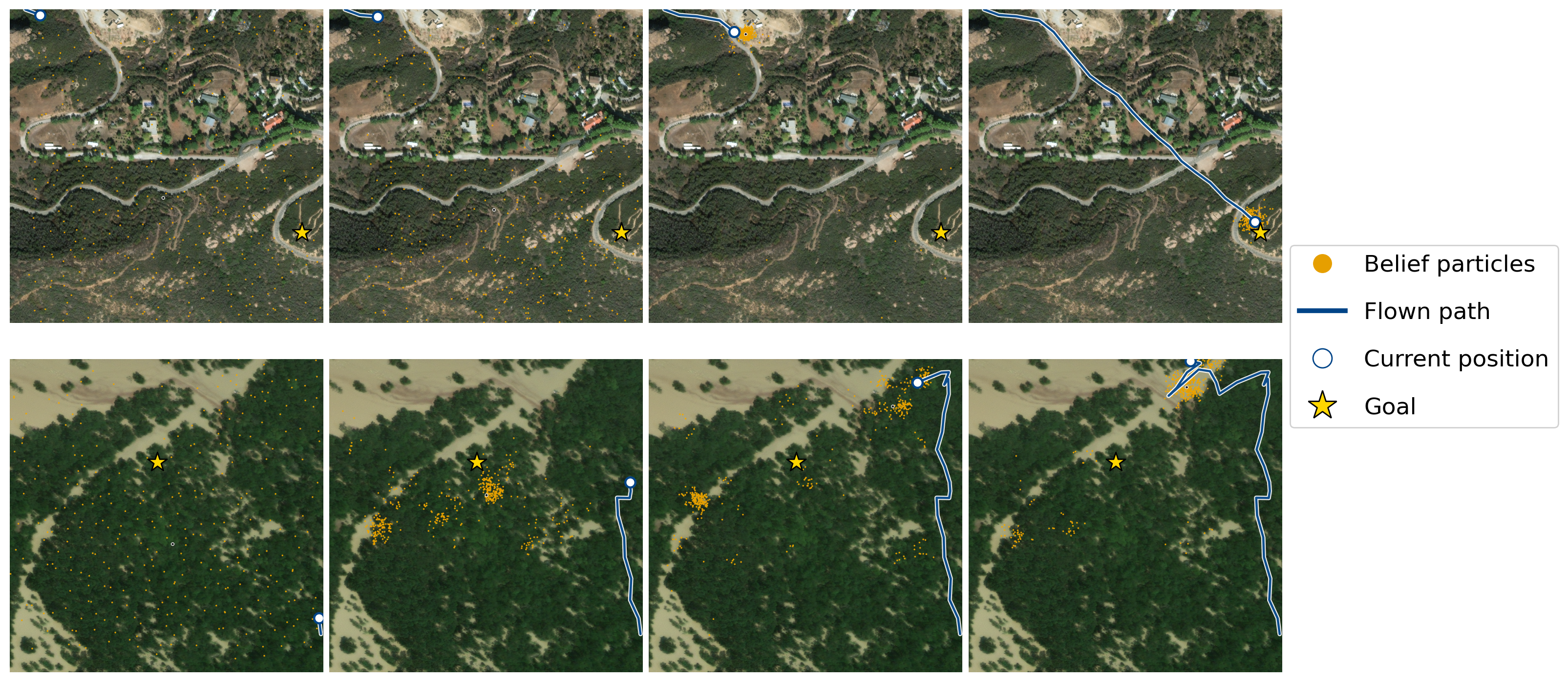}
    \caption{\textbf{When the sufficiency margin holds, and when it does not.} Each row shows four snapshots (left to right) of belief particles (orange) as the UAV flies toward the goal (star). \textbf{Top:} particles start diffuse and contract sharply once the UAV passes distinctive built-up terrain on its direct path to the goal. \textbf{Bottom:} over uniform forest canopy the observations stay ambiguous throughout. Particles keep re-clustering on visually similar but incorrect locations.}
    \label{fig:success}
\end{figure*}

\noindent\textbf{Navigation-side mechanisms.}\\*
The numbers above are for the bare pipeline, which does nothing beyond maintaining a belief and pointing at the goal, with no training, no inertial odometry, and no route planning. Mechanisms that use motion to resolve localization ambiguity, standard in active localization, do not improve on it: adding any of the four extensions of Sec.~\ref{sec:extensions} leaves performance unchanged or lowers it, analyzed in Sec.~\ref{sec:ablation}. This agrees with Thm.~\ref{thm:sufficiency}: the concentration rate depends on the margin between distinctiveness and staleness, which the controller does not alter.

\section{Further Analyses and Ablation Studies}
\label{sec:ablation}

\noindent\textbf{Protocol.}\\*
We strip the four mechanisms of Sec.~\ref{sec:extensions} cumulatively, in the order of Table~\ref{tab:ablation_results}, ending at the bare pipeline: uniform initialization, fixed likelihood weighting, arithmetic-mean pose, and direct goal-pointing control.

\noindent\textbf{Active recovery costs efficiency and does not raise success.}\\*
Removing recovery is a large efficiency win on SwissView (SCT $0.69\rightarrow0.86$, SR $0.87\rightarrow0.92$), as goal-directed motion meets distinguishing terrain without detours. On xBD the same removal moves SR from $0.60$ to $0.55$ and SCT from $0.50$ to $0.52$, both inside the confidence intervals, and the next step of the ablation (removing reinvigoration) returns SR to $0.60$ with SCT $0.56$.

\noindent\textbf{Auxiliary mechanisms are redundant.}\\*
Removing reinvigoration, geometric-median summarization, and the utility $U_t$ leaves performance within the confidence intervals in both regimes (Table~\ref{tab:ablation_results}); the filter already represents ambiguity by spreading the posterior. The bare pipeline sits within the confidence interval of the best configuration on every metric ($0.91$/$0.60$ SR): forward motion is itself an information-gathering trajectory. Geometric-median summarization is the one mechanism we would keep on a deployed system despite this. It did not hurt on either benchmark, it costs almost nothing, and Sec.~\ref{sec:theory} identifies when it matters: before $T_0$, where the guarantee is silent and an arithmetic mean can sit between modes. We report the arithmetic mean as the default only to keep the evaluated pipeline minimal.

This agrees with Cor.~\ref{cor:regimes}. Where distinctiveness exceeds staleness along the direct route, a detour buys a bounded reduction in $T_0$ but pays its length in full, as in the SCT loss of $0.86\rightarrow0.69$ on SwissView; where staleness dominates, as over uniform canopy (Fig.~\ref{fig:success}), no trajectory recovers evidence the map lacks. On xBD, recovery changes neither SR nor SCT beyond the confidence intervals, so these results do not show that it helps there. The ablations are cumulative, in the order of Table~\ref{tab:ablation_results}, so they measure each mechanism against those still active and not in isolation; a mechanism that helps only in combination would not show here. The results show that on these two benchmarks no active mechanism improves on direct goal-directed flight. They do not show that active localization is useless in general, and Cor.~\ref{cor:regimes} does not claim it. 

\section{Conclusion}
\label{sec:conclusion}

Retaining a belief over multimodal hypotheses lets a passive goal-pointing controller navigate against stale maps, where SLAM commits to a wrong fix and per-environment RL is brittle. Four standard active-localization mechanisms fail to improve on it, and the bounds say why: belief concentration is set by map staleness and terrain distinctiveness, which the route samples but cannot change. The gain available to any controller is therefore capped by how much that margin varies over the region it can reach. On both benchmarks it varies little, and the gain is correspondingly absent. Where it varies sharply the cap is loose and a detour may pay. This evaluation does not isolate that regime, and it does not close the gap from aligned orthoimagery to onboard views.

\paragraph{Limitations.} Observations are crops of aerial and satellite orthoimagery rather than onboard UAV frames, so tilt, motion blur, altitude variation, and the oblique-to-nadir domain gap are absent. Neither benchmark isolates the heterogeneous-margin regime in which an information-seeking detour should help, which is the sharpest untested prediction of Cor.~\ref{cor:regimes}. We report navigation outcomes but not single-frame retrieval accuracy, so the share of the task the filter contributes over direct descriptor matching is not separated out. The analysis assumes an obstacle-free domain and planar motion at fixed altitude.
{
    \small
    \bibliographystyle{ieeenat_fullname}
    \bibliography{main}
}

\end{document}